\documentclass[10pt,twocolumn,letterpaper]{article}
\usepackage{cvpr}
\usepackage{graphicx}
\usepackage{amsmath}
\usepackage{amssymb}
\usepackage{booktabs}
\usepackage{times}
\usepackage{microtype}
\usepackage{epsfig}
\usepackage[table,xcdraw]{xcolor}
\usepackage{caption}
\usepackage{float}
\usepackage{placeins}
\usepackage{color, colortbl}
\usepackage{stfloats}
\usepackage{enumitem}
\usepackage{tabularx}
\usepackage{xstring}
\usepackage{multirow}
\usepackage{xspace}
\usepackage{url}
\usepackage{subcaption}
\usepackage[ruled,linesnumbered]{algorithm2e}
\usepackage{algorithmic}

\newcommand{\model}{IPVR\xspace}
\newcommand{\modelFull}{Interactive Prompting Visual Reasoner}

\newcommand{\moduleAA}{\textbf{\textit{see}} module\xspace}
\newcommand{\moduleBB}{\textbf{\textit{think}} module\xspace}
\newcommand{\moduleCC}{\textbf{\textit{confirm}} module\xspace}
\newcommand{\moduleAABold}{See Module\xspace}
\newcommand{\moduleBBBold}{Think module\xspace}
\newcommand{\moduleCCBold}{Confirm Module\xspace}

\definecolor{LightGray}{gray}{0.9}
\definecolor{ForestGreen}{RGB}{34,139,34}
\definecolor{Gray}{gray}{0.9}
\SetKwInput{KwData}{Input}
\SetKwInput{KwResult}{Output}

% ...

\newcommand{\R}[1]{{%
    \textbf{%
        \ifstrequal{#1}{1}{\textcolor{red}{R#1}}{%
        \ifstrequal{#1}{2}{\textcolor{blue}{R#1}}{%
        \ifstrequal{#1}{3}{\textcolor{magenta}{R#1}}{%
        \ifstrequal{#1}{4}{\textcolor{teal}{R#1}}{%
                           \textcolor{cyan}{R#1}%
        }}}}%
    }%
}}

\usepackage{xr-hyper}

\makeatletter
\newcommand*{\addFileDependency}[1]{
  \typeout{(#1)}
  \@addtofilelist{#1}
  \IfFileExists{#1}{}{\typeout{No file #1.}}
}

\makeatother

\usepackage[pagebackref,breaklinks,colorlinks]{hyperref}
\usepackage[capitalize]{cleveref}
\crefname{section}{Sec.}{Secs.}
\crefname{table}{Table}{Tables}
\crefname{figure}{Fig.}{Figs.}

\begin{document}
%% TITLE
\title{See, Think,  Confirm: 
Interactive Prompting Between \\ Vision and Language Models for Knowledge-based Visual Reasoning}

\author{ Zhenfang Chen$^1$\thanks{indicates equal contributions} \quad
Qinhong Zhou$^{2*}$\quad
 Yikang Shen$^1$ \quad\\
 Yining Hong$^3$\quad
 Hao Zhang$^4$ \quad
 Chuang Gan$^{1,4}$ \\
$^1$MIT-IBM Watson AI Lab\quad $^2$Tsinghua University \\ \quad $^3$University of California, Los Angeles\quad $^4$UMass Amherst
}

\maketitle

\begin{abstract}
   Large pre-trained vision and language models have demonstrated remarkable capacities for various tasks. However, solving the knowledge-based visual reasoning tasks remains challenging, which requires a model to comprehensively understand image content, connect external world knowledge, and perform step-by-step reasoning to answer the questions correctly.
   To this end, we propose a novel framework named \modelFull~(\model) for few-shot knowledge-based visual reasoning.
   \model contains three stages, \textbf{see}, \textbf{think} and \textbf{confirm}. The \textbf{see} stage scans the image and grounds the visual concept candidates with a visual perception model.
   The \textbf{think} stage adopts a pre-trained large language model (LLM) to attend to the key concepts from candidates adaptively.
   It then transforms them into text context for prompting with a visual captioning model and adopts the LLM to generate the answer.
   The \textbf{confirm} stage further uses the LLM to generate the supporting rationale to the answer, verify the generated rationale with a cross-modality classifier and ensure that the rationale can infer the predicted output consistently.
   We conduct experiments on a range of  knowledge-based visual reasoning datasets. 
   We found our \model enjoys several benefits, 1). it achieves better performance than the previous few-shot learning baselines; 
   2). it enjoys the total transparency and trustworthiness of the whole reasoning process by providing  rationales for each reasoning step;
   3). it is computation-efficient compared with other fine-tuning baselines.
\end{abstract}

\section{Introduction}
\label{sec:intro}
\begin{figure}[t]
  \centering
   \includegraphics[width=\linewidth]{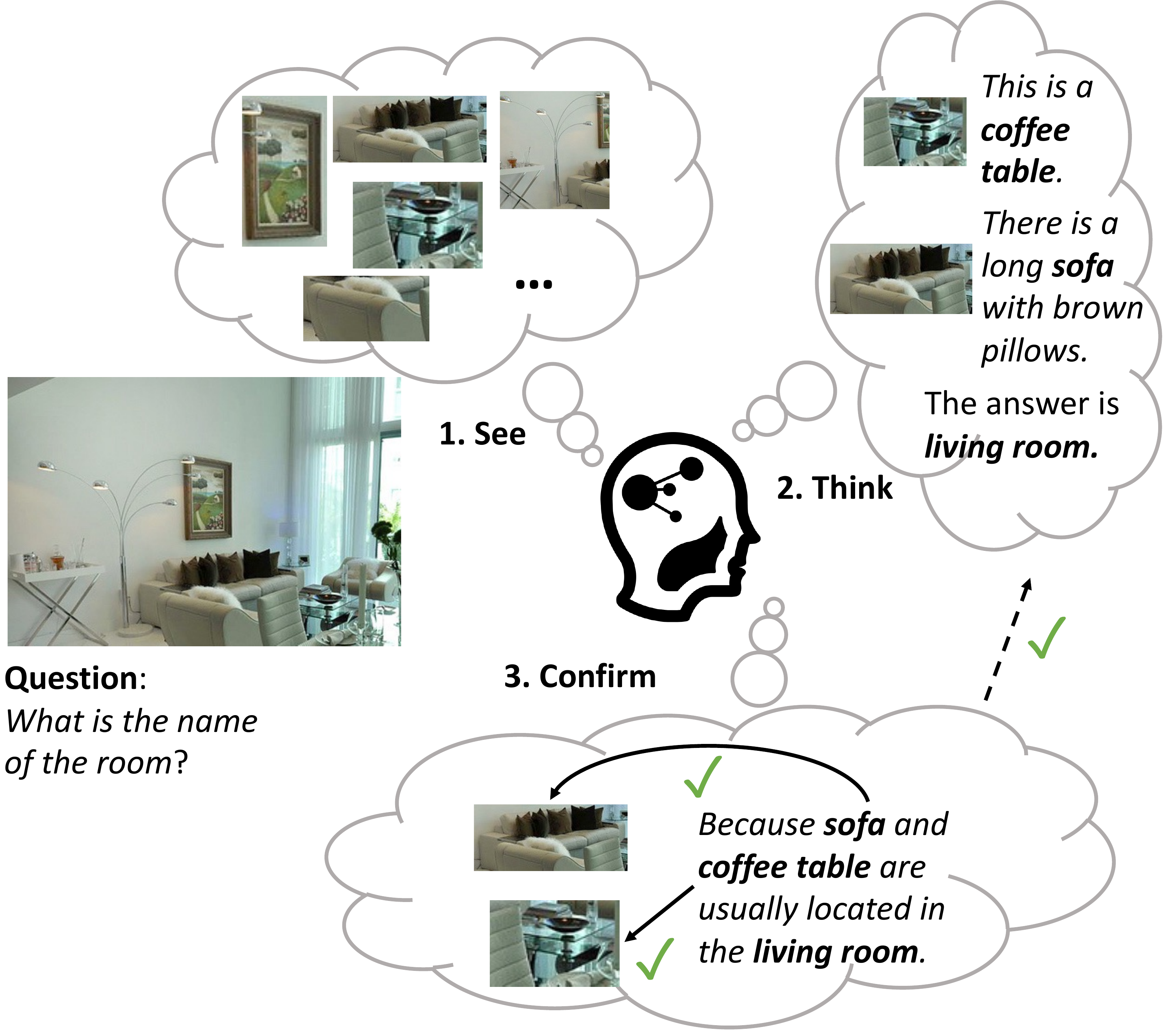}
   \caption{The human process to handle knowledge-based visual reasoning. Given an image-question pair, a human is able to \textit{see} all the objects in the image, \textit{think} of the related visual concepts to get the answer, and finally \textit{confirm} the answer is correct based on visual observation and knowledge.}
   \label{fig:teaser}
\end{figure}

We study the problem of knowledge-based visual reasoning (KB-VQA)~\cite{marino2019ok,schwenk2022okvqa}, which requires models  to recognize the image content, recall open-world knowledge, and perform logical reasoning to arrive at an answer. KB-VQA is more challenging than traditional visual question answering~\cite{antol2015vqa,balanced_vqa_v2} since the models need to understand external knowledge besides perceiving visual content. 
%\gc{since we need to ground external knowledge},
It has many real-world applications such as chatbots~\cite{shuster2022blenderbot},  assistive robots~\cite{brohan2022can}, and intelligent tutoring~\cite{anderson1985intelligent,yang2021machine}.  
As depicted in Fig~\ref{fig:teaser}, to answer the question \textit{``What is the name of the room?"}, humans need first to \textbf{\textit{see}} the image and extract visual concepts such as \textit{``frame''}, \textit{``sofa''}, and\textit{``lamp''}. We then attend to the key concepts that are semantically related to the question and \textbf{\textit{think}} that \textit{``This is a coffee table"} and \textit{``There is a long sofa with brown pillows"} to get the answer \textit{``living room"}. We can finally \textbf{\textit{confirm}} the answer is correct because \textit{``Sofa and coffee table are usually located in the living room"}.

Inspired by the success of large language models (LLMs) in few-shot natural language processing reasoning~\cite{brown2020language,zhang2022opt}, there are several works to adopt LLMs for reasoning in vision and language tasks.
The dominant approaches are mainly divided into two categories. The first category  adds additional visual perception modules to transform the visual inputs into latent inputs for LLMs, and finetunes the models with massive vision-language data~\cite{tsimpoukelli2021multimodal,alayrac2022flamingo,jin2022good}.
Although such a pipeline could achieve high performance on the downstream visual reasoning tasks, it requires a large vision-language dataset to finetune the LLM and the new visual modules for each downstream task~\cite{tsimpoukelli2021multimodal,jin2022good}, which are typically computational-intensive and time-consuming. 
The second category uses prompt-based methods for visual reasoning.
For example, PICa~\cite{yang2022empirical} translates images into captions, which can then be used as textual prompt inputs for GPT3 models~\cite{brown2020language} to answer the questions. Despite their high accuracy on knowledge-based visual question answering~\cite{marino2019ok}, their model has several limitations. 
First, the captioning processing in PICa is independent of the question's semantics,
limiting the caption to focus only on the image's general aspects instead of the question-related objects.
Second, their pipeline can not provide a step-by-step reasoning trace, leaving the question-answering a black-box process.

The key question we would like to investigate is how we can leverage the large vision and language models to achieve high accuracy for knowledge-based VQA, while maintaining transparency and efficiency of the reasoning process.
The ideas of neural-symbolic reasoning models~\cite{nsvqa,Mao2019NeuroSymbolic,yi2019clevrer,zfchen2021iclr} offer a promising step-by-step transparent visual reasoning approach. They typically transform the input question into a program consisting of a series of operations, and execute these operations with a set of network modules to get the answer. However, their success in both performance and transparency has mainly been limited to specific domains with simple visual primitive concepts (\eg, simulated physical scenes with only three shapes~\cite{johnson2017clevr}). Moreover, they often need to manually design and implement symbolic logic for each operation, limiting their applications to open-world knowledge-based visual reasoning. 

Towards more effective and interpretable complex knowledge reasoning in the visual world, we propose a novel framework named \modelFull~(\model) to mimic the human reasoning process in Fig.~\ref{fig:teaser}. Our model also contains three key modules, a \moduleAA, a \moduleBB, and a \moduleCC. Given an image-question pair, the \moduleAA first uses a scene parser to extract all the candidate visual concepts in the image. The \moduleBB adopts an LLM to select relevant visual concepts (\eg \textit{``Sofa"} in Fig.~\ref{fig:teaser}) extracted by the \moduleAA corresponding to the given task, and uses a captioning model to transform them into textual descriptions(\eg \textit{``There is a long sofa with brown pillows"}). The LLM predicts the answer to the question (\textit{``living room"}) based on the attended visual context. Moreover, we introduce a \moduleCC for rationale verification to provide more transparent and trustworthy reasoning. Specifically, we require the LLM to generate rationales (\eg \textit{``Sofa and coffee table are usually located in the living room"} for the predicted answer in Fig.~\ref{fig:teaser}). We then estimate the matching similarity between these rationales and the given visual input with a neural cross-modality classifier. Finally, the selected rationale is fed to the LLM's prompt to ensure that the rationale can infer the same output consistently.
We repeat the process of \textbf{\textit{think}} and \textbf{\textit{confirm}} iteratively until the answers from two consequent iterations are the same.

To summarize, we introduce \model, a novel modularized, interactive and iterative framework for knowledge-based visual reasoning, which is able to iteratively attend to the related visual concepts in the image and provide consistent supporting rationales for the answer prediction. \model enjoys several advantages. First, it is effective. Extensive experiments on knowledge-based benchmarks found that \model achieves better performance than previous few-shot baselines. Moreover, \model is more transparent and interpretable since it maintains the whole step-by-step reasoning trace that leads to the prediction. 
\model is also more computationally efficient compared with finetuning methods.

\section{Related Work}
\label{sec:related}

\noindent{\textbf{{Prompting Large Pre-trained Models.}}
Large language models like GPT-3~\cite{brown2020language} have popularized few-shot prompting in natural language processing (NLP), where several input-output pairs are used as context for the language model to understand the task and generate predictions for a new example. Some prompting techniques~\cite{wei2022chain,creswell2022selection,zhou2022least,marasovic2022shot} have also been developed for more effective or transparent reasoning in NLP.
Later, prompting was brought to the vision community~\cite{zhou2022learning,jia2022visual,ju2021prompting,ge2022domain,Zhou_2022_CVPR,wang2022language}. CLIP~\cite{Radford2021LearningTV} and RegionCLIP~\cite{Zhong_2022_CVPR} enable zero-shot classification and detection by replacing the class labels with natural language supervision during training. UnitedIO~\cite{lu2022unified} specifies each vision task with a language prompt to perform multi-task learning.
Differently, we aim to use interactive prompting for knowledge-based visual reasoning. It requires frequent interaction between language models and vision models, such as extracting visual concepts related to the task, recalling corresponding external knowledge, and verifying the text prediction consistent with the visual context, which has not been well studied before.

\noindent{\textbf{Large Pre-trained Models for Visual Reasoning.}}
Large pre-trained models have also been used in reasoning over vision and language~\cite{dou2022coarse,jin2022good,Wen_2021_ICCV,zeng2022socraticmodels,Gao_2022_CVPR}.
Most works~\cite{tsimpoukelli2021multimodal,lei2021less,jin2022good,zellers2022merlot,li2022blip,chen2022murag} learn large pre-trained models from massive vision-language data and finetune them for downstream tasks, which are usually extremely computationally expensive and time-consuming. For example, Flamingo~\cite{alayrac2022flamingo} needs to be finetuned on 1536 TPUv4 for 15 days with 185 million images and 182 GB of text.
It has also been observed that many of these pre-trained models (\eg,~\cite{kamath2022webly,tan2019lxmert}) contain limited open-world knowledge.
They achieve inferior performance on KB-VQA datasets, compared with models with LLMs for external knowledge~\cite{gui2021kat,yang2022empirical}.
Recently, Yang~\etal converted images into textual descriptions and treated them as prompts for LLMs, which achieves high performance on knowledge-based visual question answering. However, its text-based visual context is independent of the query question and leaves the question-answering process a black box. 

\noindent{\textbf{Knowledge-based Visual Reasoning.}}
Our work is also related to knowledge-based visual question answering (KB-VQA)~\cite{marino2019ok,schwenk2022okvqa,wang2017fvqa}, which requires both understanding the image content and retrieving external knowledge to answer the questions. Most early methods~\cite{wang2017explicit,zhu2021mucko,ding2022mukea,garderes2020conceptbert,Zhang_2022_CVPR,heo2022hypergraph,lin2022retievevqa} use deep neural networks to understand images and retrieve relevant knowledge from explicit knowledge bases. Recent methods~\cite{gui2021kat,yang2022empirical,krishna2017visual,lin2022revive} found that LLMs like GPT-3 could serve as a knowledge base and use the LLM to answer the question directly. While our model also retrieves relevant knowledge from LLMs, it provides the step-by-step reasoning process besides the final answer prediction.

\begin{figure*}[t]
  \centering
   \includegraphics[width=\linewidth]
   {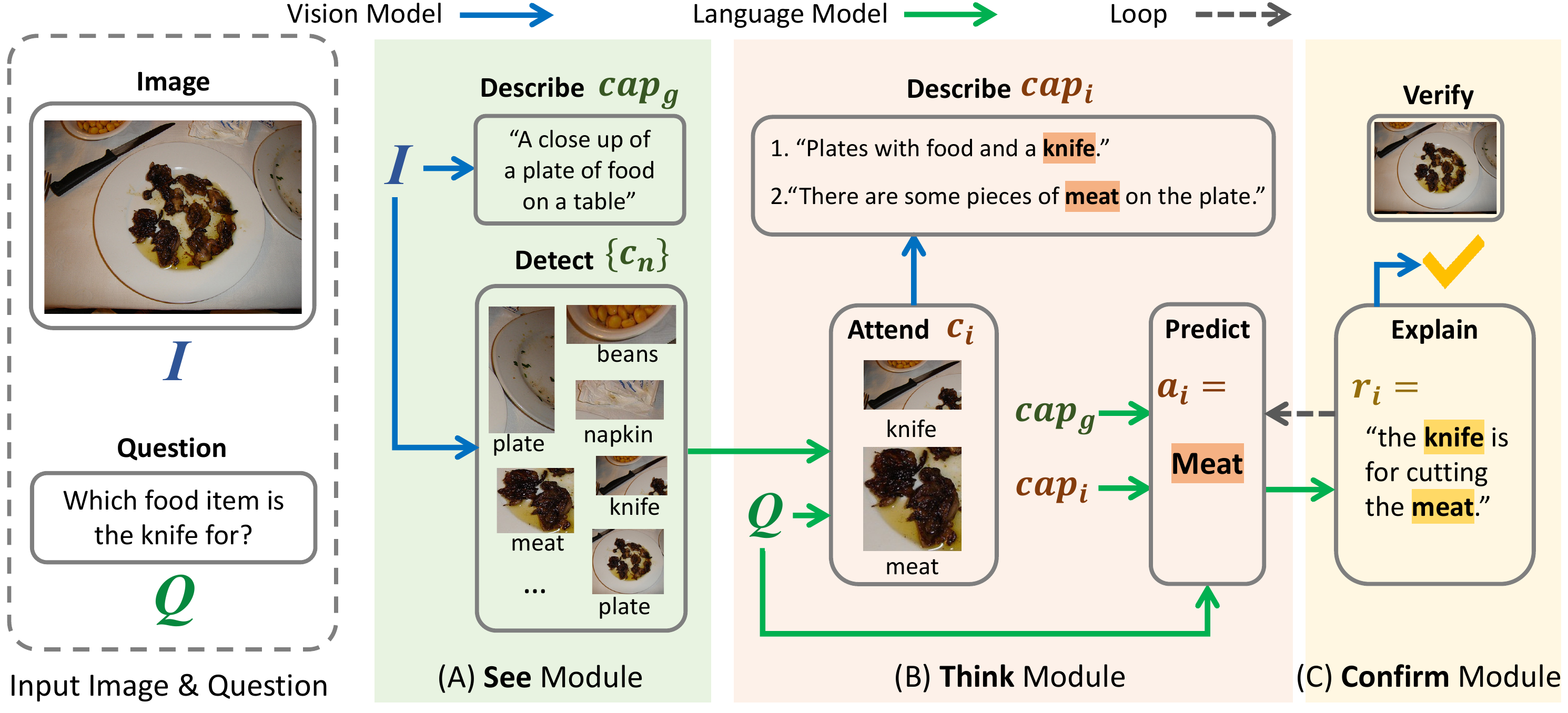}
   \vspace{-5mm}
   \caption{The framework of our \model. Given an image-question pair, we first use the \moduleAA to detect all object candidates in the image and translate the whole image into a global description. Then, the \moduleBB adopts an LLM to attend to the key visual concepts, transforms the selected concept into a language description with a captioner, and leverages the LLM to predict an answer. The \moduleCC requires the LLM to continue the generate the supporting rationale, verify whether the rationale is consistent with the image content, and ensure that the same answer can be produced when the rationale is added to the prompt in the next iteration.}
   \label{fig:frmwork}
\end{figure*}

\noindent{\textbf{Neural-Symbolic Visual Reasoning.}}
Our work could also be regarded as neural module networks~\cite{nsvqa,Mao2019NeuroSymbolic,chen2022comphy,andreas2016neural,chen2021meta,amizadeh2020neuro,ding2021dynamic,ding2022embodied}, which provides transparent step-by-step reasoning. They typically decompose the query question into a set of operations, model each operation with a network module, and iteratively execute these operations to get a transparent result. While these models are more interpretable, they either focus on simple physical scenes~\cite{johnson2017clevr,yi2019clevrer} or only achieve inferior performance on real-world datasets~\cite{hudson2019gqa,antol2015vqa}, compared with end-to-end neural network methods~\cite{li2020unicoder,chen2022pali}.
Different from them, we would like to build a system that can achieve reasonable performance on open-world knowledge-based visual reasoning while maintaining the system's interoperability. 

\section{Method}
\label{sec:method}
\subsection{Overall}
In this section, we introduce a new model called \modelFull~(\model) for knowledge-based visual reasoning, which can understand the query question, attend to key visual concepts in the image, retrieve supporting evidence, and finally get the answer in a step-by-step manner. \model consists of three modules, \textit{\textbf{see}}, \textit{\textbf{think}} and \textit{\textbf{confirm}} and run these modules in an iterative manner. As illustrated in Fig.~\ref{fig:frmwork}, given an image and a question about its content, the \moduleAA uses a scene parser \cite{han2021image} to detect all the candidate objects (concepts) in the image and represents them with their predicted class names. It also generates a global description for the whole image. Then, \moduleBB attends to the key concepts that are semantically related to the query question with an LLM~\cite{zhang2022opt} and describes them in the form of natural language with an image captioner~\cite{li2022blip}. Based on the attended visual context, the LLM predicts the answer to the question. The \moduleCC requires the LLM to continue to generate the answer's supporting rationale and verify them with a cross-modality classifier~\cite{Radford2021LearningTV}. To ensure the rationale is consistent with the answer, we add the verified supporting rationale back into the prompting context and begin a new \textit{\textbf{think}}-\textit{\textbf{confirm}} iteration. 
We iteratively generate the answer and the rationale until the answer predictions in two consequent iterations are consistent. We summarize the whole algorithm flow in Algorithm~\ref{alg:ouralg}.

Compared with existing learning-in-context methods~\cite{yang2022empirical,wei2022chain}, our framework has two advantages, effectiveness and interpretability. It is effective since it can adaptively attend to the related visual regions and output consistent answer-rationale predictions. It is interpretable as it is able to perform a step-by-step investigation of the whole reasoning process. Visualized examples of such a step-by-step reasoning process can be found in Fig.~\ref{fig:qual}.

\subsection{Model Details}
\label{sec:model}

\paragraph{\moduleAABold.}
Given a query image, we use a Faster-RCNN~\cite{ren2015faster} to detect all the object candidates in the image. Specifically, we use the detection model released by Yang~\etal~\cite{han2021image} to predict object locations and their categorical labels such as \textit{``knife"}, \textit{``plate"} and \textit{``napkin"}. It also provides a global caption for the whole image with an image captioner~\cite{li2022blip}. These visual concepts will be selected and further described in the \moduleBB to provide valuable visual context to get the answer.

\paragraph{\moduleBBBold.}
The second module of our framework is the \moduleBB, which attends to the corresponding regions in the image and transforms them into the textual description for the LLM to predict the answer.
 To achieve this goal, we use an \textit{attend}-\textit{describe}-\textit{predict} approach in the \moduleBB, as shown in Fig.~\ref{fig:frmwork}~(B).
 
 The first step is \textit{attend}, where we use prompting methods~\cite{brown2020language,chowdhery2022palm} to help the LLM to attend to the key concepts in the image that is semantically related to the query question.
 We show the prompting template for the LLM to attend to key visual concepts in Fig.~\ref{fig:prompt}~(A). We feed some input-output pairs from the training set into the LLM's prompt and ask the LLM to select based on the given context.
 
Specifically, the question is shown on the top of the template. Objects detected in the \moduleAA are represented by their category names, such as \textit{``knife''} and \textit{``beans''}, and the vocabulary of LLM output words is constrained to these category names. The LLM selects the most related object to provide further visual context to handle the query question.  As shown in Fig.~\ref{fig:frmwork}~(B), such attended concepts (\eg \textit{``meat"} and \textit{``knife"}) are important to get the answer \textit{``meat''} to the question \textit{``Which food item is the knife for?''}.

 The next step of the \moduleBB is \textit{describe}. The region in the image corresponding to the selected concept is cropped and fed to an image captioner~\cite{li2022blip} to generate a regional description for the new concept. The generated regional descriptions will be added to the LLM's prompt to provide fine-grained visual context to predict an answer. Note that a regional description for an object is usually more informative than the object class. For example, the caption of \textit{``some pieces of meat on the plate"} in Fig.~\ref{fig:frmwork} additionally describes the relationship between \textit{``meat"} and \textit{``plate"}.

 The last step of the \moduleBB is \textit{predict}. We add multiple question-answering examples from the training set to the LLM's prompt and ask it to predict an answer to the question. The answer prediction is based on the attended visual context and the rationale predicted by the \moduleCC in the previous iterations, which we will discuss in the \moduleCC. 

\begin{algorithm}[t]
\caption{Pipeline of the proposed \model}\label{alg:ouralg}
\KwData{Input Image and Question $\{\mathbf{I}, \mathbf{Q}\}$.}
\KwResult{Answer and the reasoning process $\{\mathbf{a^*}, \mathbf{R}\}$ }
\begin{algorithmic}[1]
\REQUIRE $c_n$ is the $n$-th concept detected in the image; $c_i$ and $cap_i$ are the concept and the regional caption attended at the the $i$-th iteration; $cap_g$ denotes a caption for the global image.  
$mIter$: the maximal iteration; $P_{thk}$ and $P_{con}$ are the prompt text for concept selection and question answering.
\STATE  {\color{ForestGreen} \# the \moduleAA }
\STATE $\{c_n\}_{n=1}^N \gets$ ImageParser($\mathbf{I}$)
\STATE $cap_g \gets$ GlobalCaptioner($\mathbf{I}$)
\STATE $i$ starts from 0. $a_0$ is an empty string; $P_{con, 0}$ is the in-context examples of the question, answer, and rationales; $P_{thk}$ is the in-context examples of the question and object selection. 

\REPEAT
    \STATE $i = i + 1$
    \STATE  {\color{ForestGreen} \# the \moduleBB}
    \STATE $c_i \gets$ LLMAttend$(\{c_n\}_{n=1}^N, \mathbf{Q}, P_{thk})$
    \STATE $cap_i \gets$ Captioner$(c_i, \mathbf{I})$
    \STATE $P_{con, i} = P_{con, i-1} + cap_i$
    \STATE $a_i \gets $ LLMPredict $(P_{con, i}, \mathbf{Q})$ 
    \STATE {\color{ForestGreen} \# the \moduleCC }
    \STATE $r_i \gets$ LLMConfirm $(P_{con, i}, \mathbf{Q}, a_i)$
    %\STATE \If {Verify($r_i$, Q) > thre}
    \IF{$Verify(r_i, \mathbf{I}) > thre$ }
    \STATE $P_{con, i} = P_{con, i} + r_i$ 
    \ENDIF
\UNTIL {$a_i==a_{i-1}$ \textbf{or} $i==mIter$}
\STATE $\mathbf{a^*} \gets a_i$ ; \quad $\mathbf{R} = (\{c_j, cap_j\}_{j=1}^{i}, r_i)$
\RETURN $\{\mathbf{a^*}, \mathbf{R}\}$
\end{algorithmic}
\end{algorithm}

\paragraph{\moduleCCBold.}
The last module of our framework is the \moduleCC, as shown in Fig.~\ref{fig:frmwork}~(C), which aims to generate a consistent supporting rationale for the answer prediction and verify the prediction's correctness. 
Given the few-shot example context and the interactive prompt generated by the \moduleBB, we require the LLM to continue to predict the supporting rationale after the answer prediction. A prompt example of such a \textit{question}-\textit{answer}-\textit{rationale}  template can be found in Fig.~\ref{fig:prompt}~(B). A significant problem of the LLM's prediction is that the generation procedure is a black box, and it is difficult to verify the correctness of the predicted answer and rationale. We believe a correct rationale should have two distinct features. First, the rationale should be consistent with the answer. Given the predicted supporting rationale (\textit{“The knife is for cutting the meat”} in Fig.~\ref{fig:frmwork}) for the answer (\textit{``meat"}), we should be able to predict the same answer when it is added to the context. Second, the rationale should be consistent with the visual input.

To ensure that the rationale supports the answer prediction, we feed the generated textual rationale into the LLM's prompt in the next iteration. We repeat this \textit{answer-to-rationale} and \textit{rationale-to-answer} procedure until the two consequent predicted answers are the same (Line 4 to Line 17 of the Algorithm~\ref{alg:ouralg}).
To ensure that the generated rationale is consistent with the given visual context, we use a large pre-trained cross-modality classifier~\cite{Radford2021LearningTV} to verify whether the textual rationale matches the given images or not (Line 13 to Line 15 of the Algorithm~\ref{alg:ouralg}). Only the rationale with the high matching similarity will be accepted and added to the prompt for the answer prediction in the next iteration.

\begin{figure*}[t]
  \centering
   \includegraphics[width=0.9\linewidth]{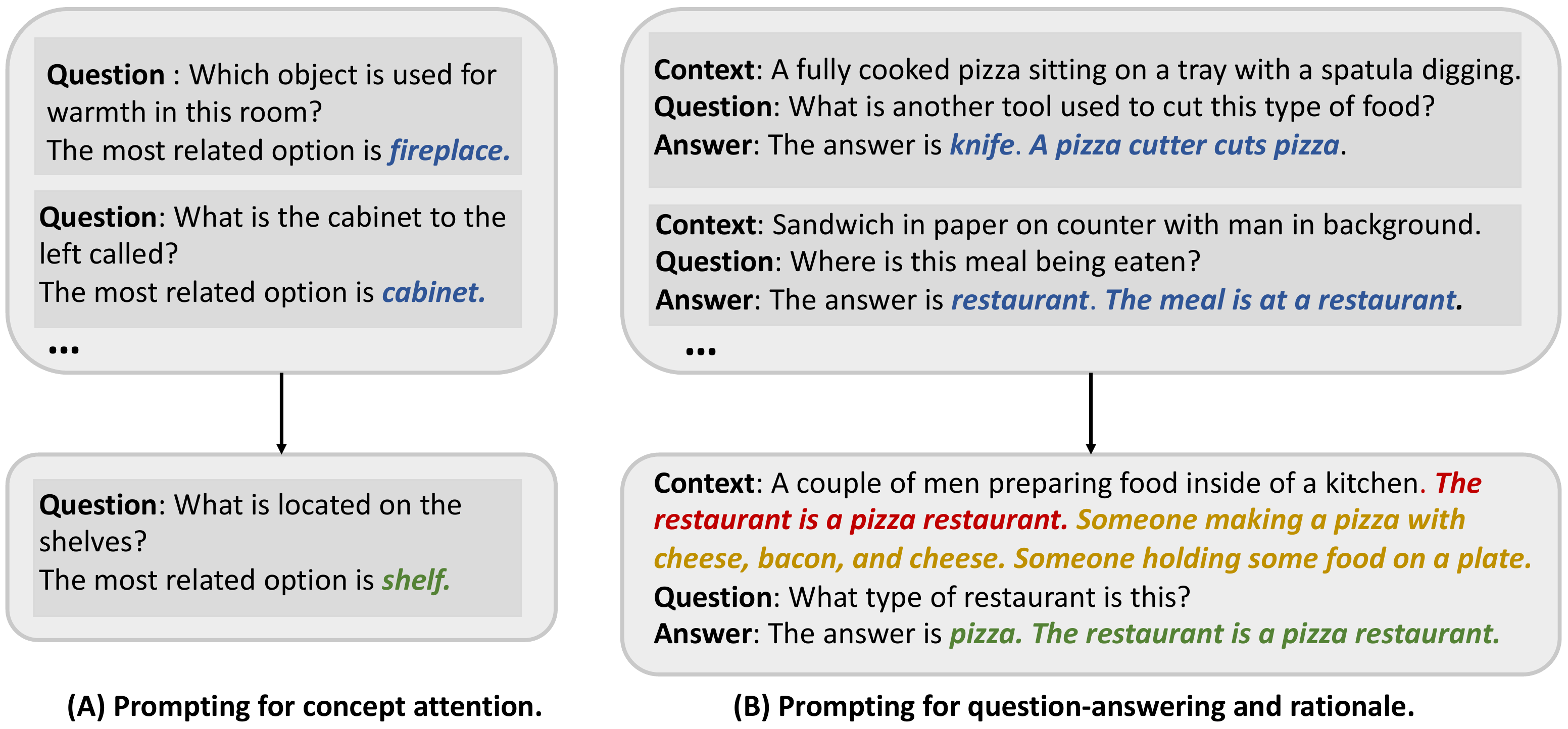}
   \vspace{-1em}
   \caption{Examples of prompting templates for visual concept attention and the full question-answer-rationale reasoning in Fig.~\ref{fig:prompt}~(A)-(B). Outputs in the in-context examples and test examples are marked with \textbf{\textcolor{blue}{blue}} and \textbf{\textcolor{ForestGreen}{green}} colors. The attentive regional captions and the rationale in the previous iterations are marked with \textbf{\textcolor{red}{red}} and \textbf{\textcolor{brown}{bronze}}~colors. We regard the most related option in the in-context examples as the option (concept) closest to the ground-truth answer by CLIP similarity.}
   \label{fig:prompt}
\end{figure*}

\section{Experiments}
\label{sec:exp}
In this section, we demonstrate the advantages of the proposed \model with extensive experiments. We first introduce our experimental setting. Then, we compare our \model with other methods on knowledge-based visual reasoning benchmarks~\cite{marino2019ok,schwenk2022okvqa} to demonstrate our method's effectiveness and interpretability. We also evaluate each module's effectiveness with an ablation study. %\zf{Experimental description for transparency.}

%\subsection{Experimental Setting.}
\noindent{\textbf{Implementation Details.}}
%\paragraph{Implementation Details.}
The effectiveness of the proposed \model relies on the interaction of several pre-trained vision models~\cite{li2022blip,han2021image,Radford2021LearningTV} and language models~\cite{zhang2022opt}. We choose the faster R-CNN model~\cite{ren2015faster} released by Han \etal~\cite{han2021image} to detect visual concepts in images, which was trained on Visual Genome~\cite{krishna2017visual} with 1,595 diverse classes. We select BLIP~\cite{li2022blip} as the regional captioning model for the attended objects. We crop the object candidate from the image and expand its bounding box 1.5 times to provide additional visual context for captioning. To make the regional captions more consistent with the question, we generate the regional caption with constrained decoding~\cite{lu2021neurologic}. We provide more implementation details in the supplementary material. We use the OPT-66B as the large pre-trained language model to prompt since it is effective, publicly available, and easy to run with 6 NVIDIA V100 PCIe 32 GB. We verify the matching similarity between the rationale and the image with the CLIP model (ViT-B/16)~\cite{Radford2021LearningTV}. We do not use GPT-3~\cite{brown2020language} for experiments due to its expensive API and limited visit frequency to the public. 

The \moduleBB of Section~\ref{sec:model} requires in-context examples to inform the LLM of the task to make predictions. We fix the number of the in-context examples to 8 since it is the largest number we could efficiently run on our hardware configuration. Following Yang~\etal~\cite{yang2022empirical}, we prompt the LLM with in-context example selection and multi-query ensemble. For in-context examples, we select the examples most similar to the current image-question pair in training set with their clip features. For multi-query ensemble, we feed our models and the baselines 5 times and select the one with the highest log-probability as previous methods~\cite{yang2022empirical,chen2021evaluating} except the aligned baseline models in Table~\ref{tab:cost}, where we ensemble 14 times for baselines to make them have similar computation cost as our model. 

\noindent{\textbf{Datasets and Evaluation Metric.}}
We evaluate our models on standard KB-VQA benchmarks, OK-VQA~\cite{marino2019ok} and A-OKVQA~\cite{schwenk2022okvqa}.
OK-VQA is the most popular knowledge-based question-answering dataset with 14,055 image-question pairs, asking questions of diverse knowledge such as transportation, food, and weather.
A-OKVQA is the current largest KB-VQA dataset, which not only asks knowledge-related questions but also provides supporting rationales, making it a better testing bed for step-by-step reasoning.
We do not conduct experiments on other KB-VQA datasets like F-VQA~\cite{wang2017fvqa} and KB-VQA~\cite{wang2017explicit}, since they assume the question knowledge could be retrieved from pre-defined knowledge bases (\eg ConceptNet~\cite{liu2004conceptnet} and Wikipedia). Pre-defined knowledge makes them less practical and not representative enough of the complex real-world knowledge visual reasoning. 
We adopt the widely-used soft accuracy~\cite{balanced_vqa_v2} as the evaluation metric.

\noindent\textbf{Baselines.}
We mainly compare our methods with two strong learning-in-context baselines, PICa~\cite{yang2022empirical} and Chain-of-Thought (CoT)~\cite{wei2022chain}.
\vspace{-1mm}
\begin{itemize}
[align=right,itemindent=0em,labelsep=2pt,labelwidth=1em,leftmargin=*,itemsep=0em]
\item \textbf{PICa}. PICa with GPT-3 is the current state-of-the-art few-shot model on OK-VQA, which prompts the LLM with only the image caption and object tags.
\item \textbf{CoT}. CoT is a popular few-shot prompting method, which asks the model to perform step-by-step reasoning to solve the task rather than directly output the answer. We implement CoT by asking the LLM to generate the reasoning step (rationale) first and then predict the answer.
\end{itemize}
\vspace{-2mm}
We carefully implement these two methods with the OPT-66B model for a fair comparison. We also include other fully-supervised methods in the tables for performance reference. Prompt examples of the baseline methods can be found in the supplementary. 

\subsection{Compared with Other Methods.}
\begin{table}[t]
    \centering
    \begin{tabular}{l|ccc}
        \toprule
        %\multirow{2}{*}{Methods} & \multicolumn{2}{c}{OK-VQA} & AOK-VQA  \\
        %&  & Val & Test & Accuracy \\        
        \multirow{2}{*}{Methods} & \multicolumn{2}{c}{A-OKVQA} & OK-VQA   \\
        & Val & Test & Test   \\
        \midrule

        %KRISP~\cite{marino2021krisp} & & & 38.35 \\
        MAVEx~\cite{wu2022multi} & - & - & 41.37 \\
        UnifER~\cite{guo2022unified} & - & - & 42.13 \\       
        Pythia~\cite{pythia18arxiv} & 25.2 & 21.9 & -\\
        ViLBERT~\cite{lu2019vilbert} & 30.6 & 25.9 & -\\
        LXMERT~\cite{tan2019lxmert}  & 30.7 & 25.9 & -\\
        KRISP~\cite{marino2021krisp} & 33.7 & 27.1 & 38.4\\
        PICa$^\star$~\cite{yang2022empirical}-GPT-3~\cite{brown2020language} & - & -& 48.0 \\
        GPV-2~\cite{kamath2022webly} & \textbf{48.6} & \textbf{40.7} & -\\
        KAT~\cite{gui2021kat}-GPT-3~\cite{brown2020language} & - & - & \textbf{54.4} \\
        \midrule
        CoT$^\star$~\cite{wei2022chain} & 41.5 & 43.7  &  38.1$^{\dagger}$ \\
        PICa$^\star$~\cite{yang2022empirical} & 42.4  & 43.8 & 42.9 \\
        \rowcolor{Gray}
        Ours$^\star$ & ~\textbf{46.4} &  \textbf{46.0} & \textbf{44.6}$^\ddagger$ \\
        %CoT~\cite{wei2022chain}-from-AOK & & & 38.13  \\
        %Ours-from-AOK & & & \textbf{38.94} \\
        %\midrule
        %PICa~\cite{yang2022empirical} & & & 42.94  \\
        %Ours w/o rationales & &  & \textbf{44.63} \\
        \bottomrule
    \end{tabular}
    \caption{Performance comparison of our model and other baselines on A-OKVQA and OK-VQA datasets. $^\star$ denotes learning-in-context methods. $\ddagger$ denotes our model's \moduleCC is removed since there are no available examples with rationales in OK-VQA.
    $^{\dagger}$ denotes in-context examples with the rationales for CoT are from A-OKVQA dataset. Our model performs better than baselines.}
    \label{tab:aokok}
\end{table}
\noindent{\textbf{{Quantitative Results.}}
We compare our \model with baselines on the validation and test sets of the A-OKVQA dataset in Table~\ref{tab:aokok}. Few-shot learning-in-context methods are marked with $\star$ in the table, and our method is marked with gray background for easy reference.
According to the results, we have the following observations. 
First, we have constant gains compared with the learning-in-context baselines, PICa and CoT and even achieve better performance than the previous full-supervised pre-trained method GPV-2 on the test split of A-OKVQA. These gains show our method's effectiveness in answer predictions. We have also noticed that fully-supervised methods have a more significant performance disparity between the validation and test splits than the learning-in-context methods. For example, GPV-2 has an accuracy drop of 7.9, while our method only drops 0.4. We believe the reason is that the validation set has frequently been evaluated in fully-supervised methods in different epochs and is somewhat overfitted. 

We also evaluated our method in the OK-VQA in Table~\ref{tab:aokok}. The training set of the OK-VQA does not provide rationales for reasoning, which is needed in CoT and our method.
For our method,  we develop a variant that only uses examples in OK-VQA as prompting for the \moduleBB without rationale reasoning. The model stops when the predicted answers from two consecutive iterations become the same.
We observe that our model performs better than the baseline methods, PICa and CoT, which is consistent with our finding in the A-OKVQA dataset.
We can also see that the state-of-the-art methods (KAT-GPT-3 and PICa-GPT-3) rely on much more powerful LLM, GPT-3 (175B), to achieve great performance. In contrast, our method achieves reasonable performance by integrating the publicly-available OPT-66B LLM.

\noindent{\textbf{Qualitative Results.}} 
The step-by-step reasoning nature of our \model provides better transparency and interoperability, which makes it easy to understand how our model works.
We provide a qualitative comparison with baselines in Fig.~\ref{fig:qual}.
The $\rightarrow$ in Fig.~\ref{fig:qual} shows the flows of our method's reasoning process to get related visual concepts, regional descriptions, and the supporting rationale. 

Compared with PICa, our method can adaptively attend to key visual concepts (\eg \textit{``ball''} and \textit{``tennis court"} in Fig.~\ref{fig:qual}) in the image that are semantically important to the question (\eg \textit{What is the fence meant to block?}) and describe that in the form of the natural language (\eg \textit{``Someone has a tennis racket and is about to hit the ball''}) to get the answer.
Besides, as shown in the \textit{explain} column of Fig.~\ref{fig:qual}, our method could generate better supporting rationale (\textit{``The wall is used for displaying art"}), which matches with the visual context in images better.
In contrast, the CoT might generate a rationale inconsistent with visual context (\textit{``The wall is used for a sofa."}), which leads to a wrong answer.

\begin{figure*}[t]
  \centering
   \includegraphics[width=\linewidth]{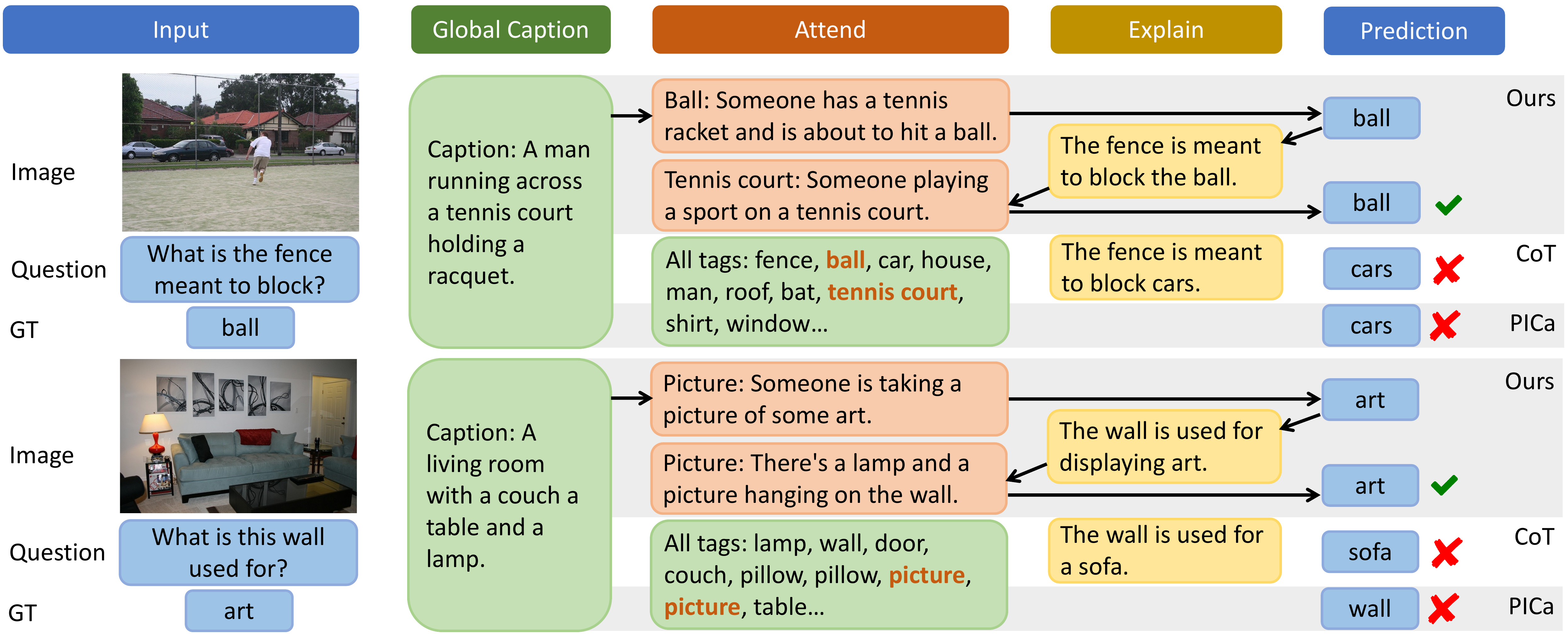}
   \caption{Qualitative results of our \model and baselines. Besides better answer accuracy, our method also enjoys better transparency by providing a step-by-step reasoning trace with related visual concepts, regional descriptions, and a supporting rationale. $\rightarrow$ denotes our reasoning flow. 
   }
   \label{fig:qual}
\end{figure*}

\noindent{\textbf{Rationale Evaluation.}}
To further evaluate the reasoning process of our method, we compare the quality of rationales between our method and CoT in Table~\ref{tab:aok-rationale} on the validation set of A-OKVQA, where the rationales are publicly available. We use widely-used BLEU scores and CLIP sentence similarity as the metrics. We measure BLEU calculated by \textit{multi-bleu.perl}~\footnote{\url{https://github.com/moses-smt/mosesdecoder/blob/master/scripts/generic/multi-bleu.perl}} and averaged cosine similarity of sentence representations calculated by CLIP (ViT-B/16)~\cite{Radford2021LearningTV} text encoder. Both BLEU and similarity results show that the rationales generated by our method are closer to the ground truth than the rationales generated by CoT.
\begin{table}[t]
    \centering
    \begin{tabular}{l|cccccc}
        \toprule
        Methods & BLEU & Sentence Similarity \\
        \midrule
        CoT & 14.19 & 80.92   \\
        \rowcolor{Gray}
        Ours & ~\textbf{14.34} & \textbf{81.22}  \\
        %\midrule
        %CoT~\cite{wei2022chain}   \\
        %Pica~\cite{yang2022empirical} \\
        %Ours & & & \\
        \bottomrule
    \end{tabular}
    \caption{Rationale performance comparison of our model and CoT baseline on A-OKVQA validation set. The rationales generated by our \model are closer to ground
truth than those generated by CoT.}
    \label{tab:aok-rationale}
\end{table}

\begin{table}[t]
    \centering
    \begin{tabular}{l|cc}
        \toprule
        Methods & \ A-OKVQA & \ OK-VQA \\
        \midrule
        PICa & 42.40 & 42.94 \\
        PICa-aligned & 41.90 & 42.84 \\
        \midrule
        CoT & 41.53 & 38.13 \\
        CoT-aligned & 42.10 & 38.15  \\
        \midrule
        \rowcolor{Gray}
        Ours & \textbf{46.41} & \textbf{44.62} \\
        \bottomrule
    \end{tabular}
    \caption{Analysis of computational cost on A-OKVQA validation set and OK-VQA set. Our method still outperforms PICa and CoT after considering the computational cost issue.}. 
    \label{tab:cost}
\end{table}

\noindent{\textbf{Computation Analysis.}}
Although we have suggested that our method is more efficient than finetuning methods (\eg ~\cite{alayrac2022flamingo} and ~\cite{jin2022good}), it isn't entirely obvious that this is still the case when compared with other in-context learning methods. After all, the step-by-step manner of \model requires more queries to large models. To better understand the efficiency of our method, we make baselines have similar computational costs as our model. Specifically, we increase the number of queries to ensemble $k$ (5 in all the experiments except PICa-aligned and CoT-aligned below) for PICa and CoT, and make their overall amount of queries to LLM the same as our method. Considering it takes $2.27$ rounds on average for our method to get final answers and one additional query in each round introduced by \moduleBB, the mean amount of queries is $13.62$ for our method. Therefore, the cost-aligned PICa and CoT have an ensemble amount $k=14$ for each sample, denoted as PICa-aligned and CoT-aligned.

We compare their results with our method in Table~\ref{tab:cost}, which indicates that our method still outperforms PICa and CoT significantly after considering the computational cost issue. We also notice a slight performance drop for PICa when its ensemble amount $k$ increases. We suggest that this drop is related to the in-context sample selection method, which chooses the top $n*k$ questions with the highest similarities, where $n$ is the number of the in-context examples (8 in our case). When $k$ increases, the in-context questions of PICa and CoT become more irrelevant to the current question.

\noindent{\textbf{Failure Cases.}}
One distinct advantage of our method is transparency and interoperability, enabling us to analyze how the model fails. In Fig.~\ref{fig:failure}, we provide two typical failure cases of our method from the A-OKVQA validation set. Our method fails at the \textit{predict} step in the first case. Although the information from context is enough to find the answer, OPT-66B fails to predict it correctly. In the second case, our method fails at the \textit{detect} step. The key object name (``\textit{pacifier}'') is not in the object list provided by R-CNN. The failure cases indicate that our method can perform better when more powerful vision and language models are provided.

\begin{figure}[t]
  \centering
   \includegraphics[width=\linewidth]{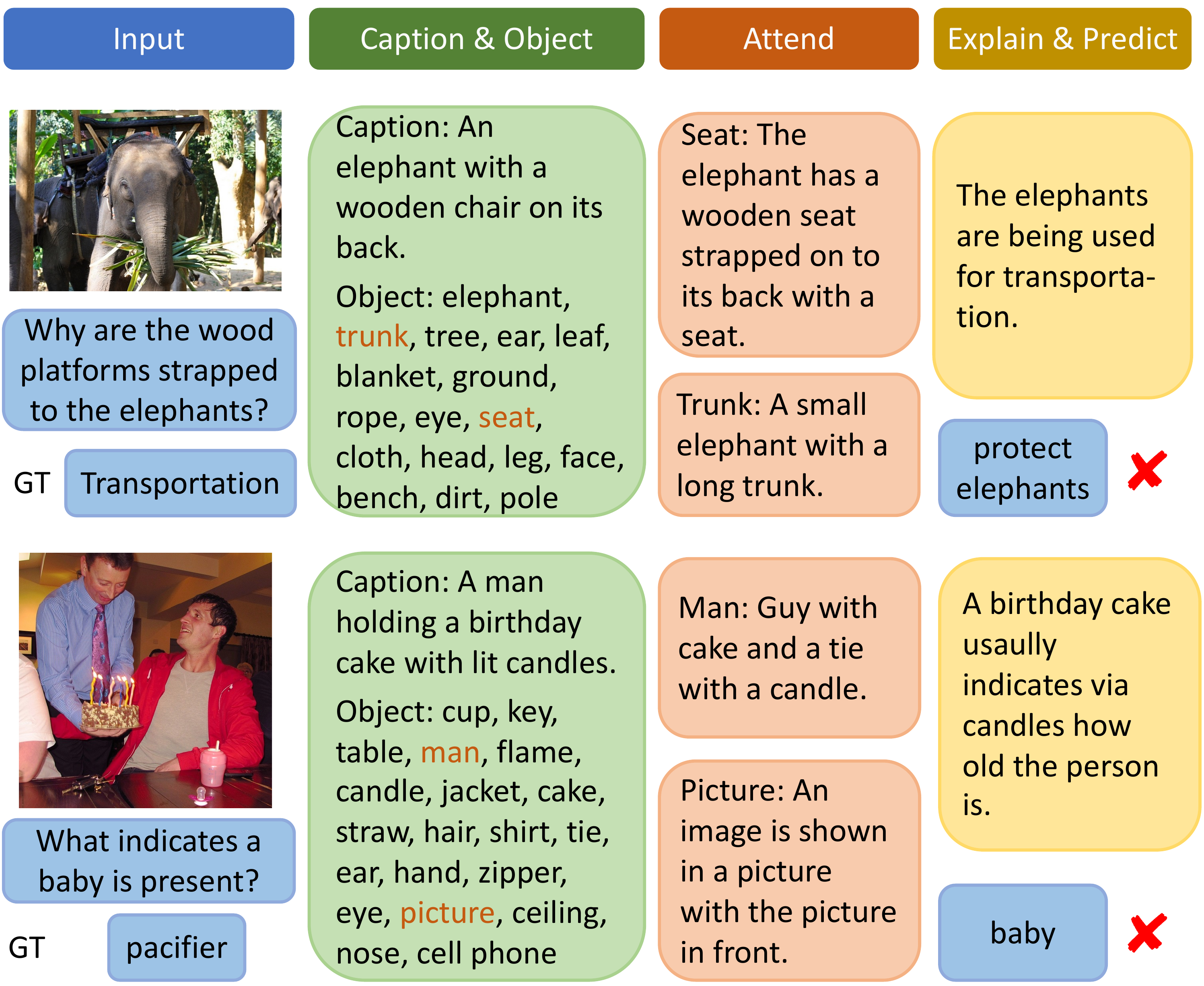}
   \caption{Failure cases on A-OKVQA validation set. Our model will collapse when the scene parser fails to \textit{detect} the key concept (\textit{``pacifier"} at the bottom)  or when the LLM fails to \textit{predict} the answer (prediction on the top) based on the correct visual context.}%\zf{Fonts are too small?}}
   \label{fig:failure}
\end{figure}

\noindent{\textbf{Ablation Study.}}
We conduct a series of ablation studies to answer the following questions. \textbf{Q1}: Is the \textit{attend} and \textit{describe} components in the \moduleBB necessary to achieve good performance?
\textbf{Q2}: Does the rationale reasoning in \moduleCC improve the answer accuracy? \textbf{Q3}: Is it necessary to verify the rationale's consistency with the input image? 
As shown in Fig.~\ref{fig:qual}, \textit{Full} denotes our models with all the component integration. \textit{``w/o attend"} denotes the \model model without the \textit{attend} and \textit{describe} components in the \moduleBB. \textit{``w/o rationale"} shows the model without generating the rationale in the \moduleBB. \textit{``w/o verify"} denotes the model without CLIP verification in the \moduleCC and simply adds the rationales back into the next answer prediction step. 

\begin{figure}[t]
  \centering
   \includegraphics[width=0.75\linewidth]{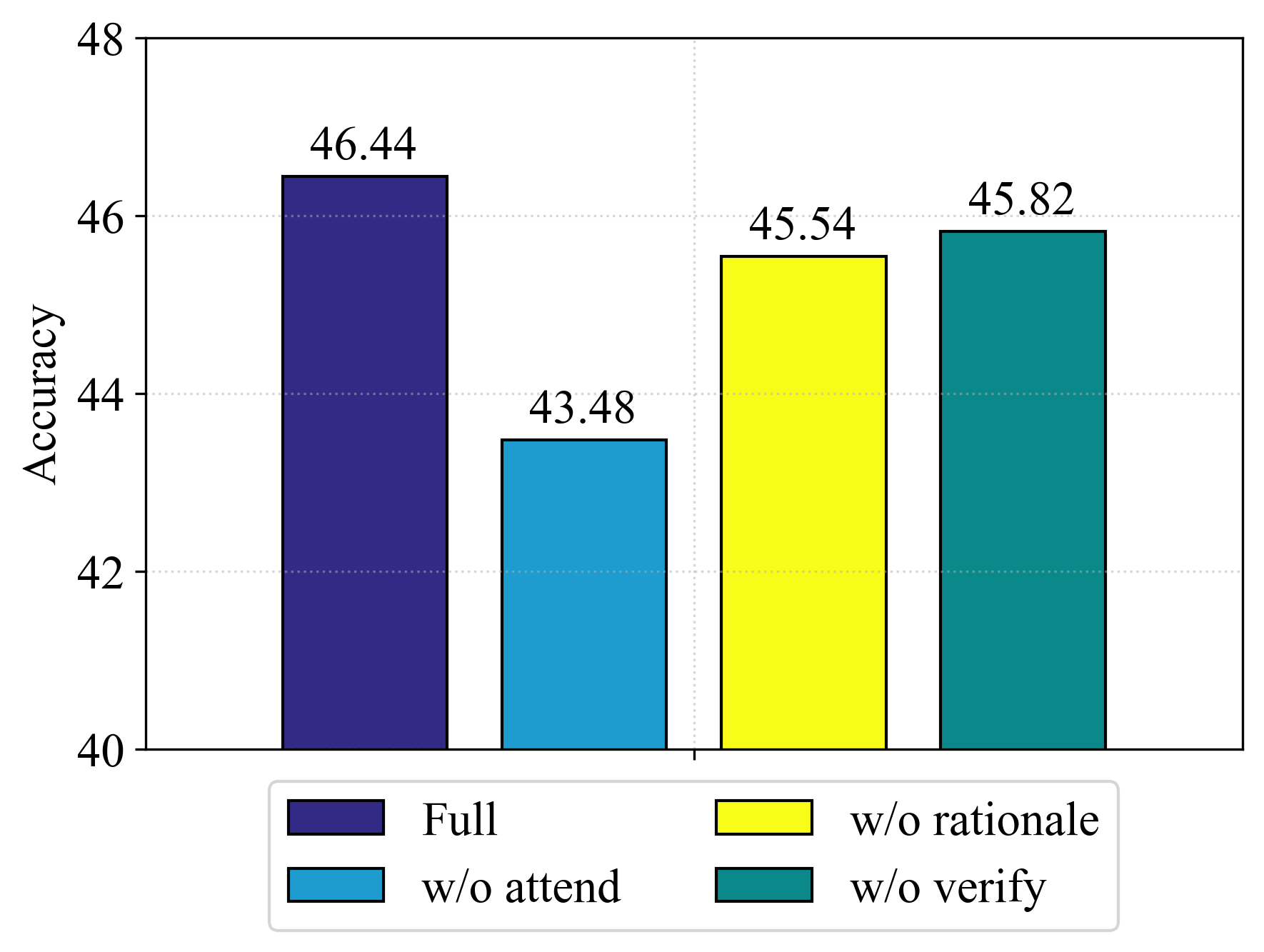}
   \vspace{-3mm}
   \caption{Ablation study on A-OKVQA validation set. Each ablated component has its contribution to the overall performance.}
   \label{fig:abs}
\end{figure}

We find that the performance will drop without any ablated component, showing each component has its contribution to the overall performance. We observe that the \textit{``attend"} module has the most significant effect on the model's overall performance, where we think the reason is that the \textit{``attend"} component has provided essential visual context for the LLM to infer the correct answer as shown in the examples in Figure~\ref{fig:qual} (answering \textbf{Q1}).
The \textit{``w/o rationale"} ablation shows that adding the reasoning rationale into the model not only increases the model's transparency but also has positive effects on the answer prediction (answering \textbf{Q2}).
From the result of \textit{w/o ``verify"} ablation, we can observe that further gain could be achieved if we verify the matching similarity between the rationale and the input image (answering \textbf{Q3}).

\section{Conclusion}
\label{sec:conclusion}
In this paper, we develop a novel model named \modelFull~(\model) for knowledge-based visual reasoning, which requires the model to understand both the image content and external knowledge to answer the query question. 
\model can adaptively focus on the related visual concepts in the image, transform them into natural language and provide consistent rationales to support the answer prediction.
Compared with existing prompting methods, it not only achieves better performance but also maintains high transparency by keeping the whole trace of each reasoning step.
We hope that our \model can motivate more future research on interactive prompting between pre-trained models of different modalities for building more effective and interpretable visual commonsense reasoning systems.

{\small
\bibliographystyle{ieee_fullname}
\bibliography{references}
}

%\ifarxiv 
\clearpage 
%\input{12_appendix}
%\fi
\appendix
\label{sec:appendix}
%\input{_constants}
%\review % \review OR \cameraready

%\documentclass[10pt,twocolumn,letterpaper]{article}
%%
%{
%    \hypersetup{linkcolor=black}
%    \tableofcontents
%}
%\clearpage

%\renewcommand{\thetable}{A\arabic{table}}
%\renewcommand{\thefigure}{A\arabic{figure}}
%\renewcommand{\thesubsection}{A\arabic{subsection}}
%\renewcommand{\thesection}{A\arabic{section}}

\section{Overview.}
In this supplementary material, we back up our claims by supplementing the main paper with more implementation details (Section~\ref{sup:imp}), more quantitative performance analysis (Section~\ref{supsec:quan}), and more qualitative visualization (Section~\ref{supsec:qual}).

\section{More Implementation Details.}
\label{sup:imp}
%\paragraph{Parameter Setting.}
\noindent{\textbf{Parameter Setting.}}
As shown in Algorithm 1 of the main paper, we stop iteratively attending to the new concept and predicting a new answer when the current predicted answer $a_i$ equals to the last predicted answer $a_{i-1}$, or it reaches the maximal iteration number $mIter$. In our implementation, we simply set $mIter$ to 5. We find that our method often converges to the same answer and jumps out of the loop after the second or the third iteration. It takes 2.27 iterations on average to get the consistent answer prediction. We add the generated rationale back into the prompt context when the cosine similarity between the generated rationale and the input image is larger than a threshold $thre$. We estimate the cross-modality similarity with CLIP~\cite{Radford2021LearningTV}. We simply set the $thre$ to be 0 and found it performs well. Our code and data will be released publicly upon acceptance.
% Our model requires regional captions for comparison. maybe a baseline model with sg_scene graphs
%\paragraph{Prompting templates.} 

\noindent{\textbf{Prompting templates}}
We also provide prompting template examples of baseline methods PICa~\cite{yang2022empirical} and CoT~\cite{wei2022chain} in Fig.~\ref{fig:pica_template} and Fig.~\ref{fig:cot_template}, respectively. Comparing the baselines' examples in Fig.~\ref{fig:pica_template}-\ref{fig:cot_template} with our method's example in Fig. 3 of the main paper, we can find that our method's prompt could provide additional visual context and rationales predicted in previous iterations to the large language model for a better and more consistent prediction.

\noindent{\textbf{Regional Captions.}
%\paragraph{Regional Captions.} 
Our model requires a regional captioning model to extract captions for each visual region (concept) selected by the \textit{attend} stage of \moduleBB. We use the pre-trained model released by Li \etal~\cite{li2022blip} to extract regional captions.
To provide additional visual context for captioning, we expand the width and length of the candidate object bounding box by 1.5 times. To make the generated caption focus more on the query question and the provided concept, we use the guided decoding strategy introduced in~\cite{lu2021neurologic} for guided captioning decoding. Specifically, we define the lookahead heuristics as the cosine similarity between the generated captions and the question estimated by a RoBERTa model~\cite{debruyn2021mfaq,conneau-etal-2020-unsupervised}~\footnote{\url{https://huggingface.co/clips/mfaq}}. We provide an ablation study between the performance of such guided captions and the ``captions" generated by naive scene graph transformation (\eg ``a silver car" for ``car" with the ``silver" attribute) in table~\ref{sup:cap}. As could be seen in table~\ref{sup:cap}, such guided caption decoder (\textbf{Ours-Guided}) provides captions of higher quality for knowledge-based visual question answering.
\begin{table}[t]
    \centering
    \begin{tabular}{l|cc}
        \toprule
        Methods & \ A-OKVQA &  \\
        \midrule
        Ours-SG &  44.7  \\
        %Ours-BLIP \\
        Ours-Guided & \textbf{46.4}  \\
        \bottomrule
    \end{tabular}
    \caption{Regional Caption generation comparison on A-OKVQA validation set with the direct-answer setting. Our method achieves consistent improvements compared with baselines.} 
    \label{sup:cap}
\end{table}

\begin{table}[t]
    \centering
    \begin{tabular}{l|cccccc}
        \toprule
        \multirow{2}{*}{Methods} & \multicolumn{2}{c}{Multiple-Choice}  \\
        &  val & test  \\
        \midrule
        Pythia~\cite{pythia18arxiv}  & 49.0 & 40.1 \\
        ViLBERT~\cite{lu2019vilbert} & 49.1 & 41.5 \\
        LXMERT~\cite{tan2019lxmert}  & 51.4 & 41.6 \\
        KRISP~\cite{marino2021krisp} & 51.9 & 42.2 & \\
        GPV-2~\cite{kamath2022webly} & \textbf{60.3} & \textbf{53.7}  \\
        \midrule
        CoT~\cite{wei2022chain}  & 48.1  & 45.6 \\
        Pica~\cite{yang2022empirical} & 46.1 & 44.2 \\
        \rowcolor{Gray}
        Ours (OPT-66B) & \textbf{48.6} & \textbf{47.6} \\
        \midrule
        \rowcolor{Gray}
        Ours (GPT-3) w/o ensemble & \textbf{58.7} & \textbf{57.5} \\
        \bottomrule
    \end{tabular}
    \caption{Performance comparison of our model and other baselines on the A-OKVQA dataset with the multiple-choice setting. Our method with GPT-3 (175B) as the language model achieves better performance than previous fully-supervised methods like GPV-2~\cite{kamath2022webly}.}
    \label{tab:mc}
\end{table}

\section{More Analysis of the Proposed Model.}
\subsection{More Quantitative Analysis.}
\label{supsec:quan}

In this section, we provide more experimental analysis of our method and baselines. Besides the direct answer that generates free-form answer predictions for questions, AOK-VQA dataset also provides a simplified and less challenging setting, where the model only needs to select an option candidate to output the answer prediction. We also compare our method with baselines in this setting in table~\ref{tab:mc}. We add the option candidates into the context of the prompt to inform the LLM what to select. Based on the experimental result, we have the following observation, our method still outperforms the baselines in this simplified setting, showing the effectiveness of our method. We also find that GPV-2 performs better than our method with the OPT-66B LLM~\cite{zhang2022opt}, which is inconsistent with our observation in table 1 of the main paper on the more challenging direct answer situations.

We analyze the failure cases of our method and baselines. We found that the released OPT-66B model has weak capabilities for selecting a correct answer from the candidates. To further prove our investigation, we replace the OPT-66B LLM of our model with the current state-of-the-art LLM engine (GPT-3 175B). We found that our model has an accuracy of 57.5 on the test split of A-OKVQA, achieving better performance than previous fully-supervised methods like GPV-2~\cite{kamath2022webly}. This shows the potential of our model that performance could be further improved with a stronger LLM like GPT-3 175B~\cite{wang2022language} or bloomz~\cite{muennighoff2022crosslingual}. Due to GPT-3 175B's expensive API cost and limitation on query frequency, we could not conduct extensive experiments with multi-query ensemble~\cite{yang2022empirical}. We expect better performance could be achieved with multi-query ensemble. We noticed that fully-supervised methods  like (GPV-2 and KRISP) have a more significant performance disparity between the validation and test splits than the learning-in-context methods, which is consistent with what we have observed in the direct-answer setting of the A-OKVQA dataset in the main paper. KRISP and GPV-2 have an accuracy drop of 9.7 and 6.6, respectively, while our model only drops 1.2. Since the validation set has frequently been evaluated in fully-supervised methods in different epochs, it may suffer from overfitting, while learning-in-context methods do not have such overfitting problems. 

\subsection{More Qualitative Analysis.}
\label{supsec:qual}

%\paragraph{Qualitative Examples.}
\noindent{\textbf{Qualitative Examples.}}
We provide more examples of our methods in Fig.~\ref{sup:qual}. As observed in the figure, we can further confirm that our method is able to additionally attend to the visual context that are semantically related to the task (\eg \textit{``pizza"}) and \textit{``food"} in the second row of Fig.~\ref{sup:qual}. Our method can also generate consistent rationales like (\eg \textit{``The restaurant is a pizza restaurant."}) for the answer prediction. Moreover, our method can provide a modularized, step-by-step investigation of the answer prediction, leading to better model transparency and interpretability. %\zf{The first case of visualization fails to attend relevant information}

\noindent{\textbf{Failure Case Analysis.}}
%\paragraph{Failure Case Analysis.}
The step-by-step reasoning trace of our model also provides intermediate results for us to analyze how the proposed \model fails and suggests further directions on how to improve the model's performance. For example, in the first row of the Fig.~\ref{sup:fail}, we found that our \model fails because our OPT-66B LLM fails to connect the \textit{``colorful balloons"} and the \textit{``lgbtq"} social movement. It shows that a better LLM with broader open-world knowledge and stronger reasoning ability  could further our method achieves better performance. The failure case in the second row of Fig.~\ref{sup:fail} shows that our model fails when the vision models fail to provide essential visual context (\ie the shape of the \textit{donut}). This indicates that our model could be further improved with models that have stronger perception abilities. 

\begin{figure}[t]
    \centering
    \includegraphics[width=0.8\linewidth]{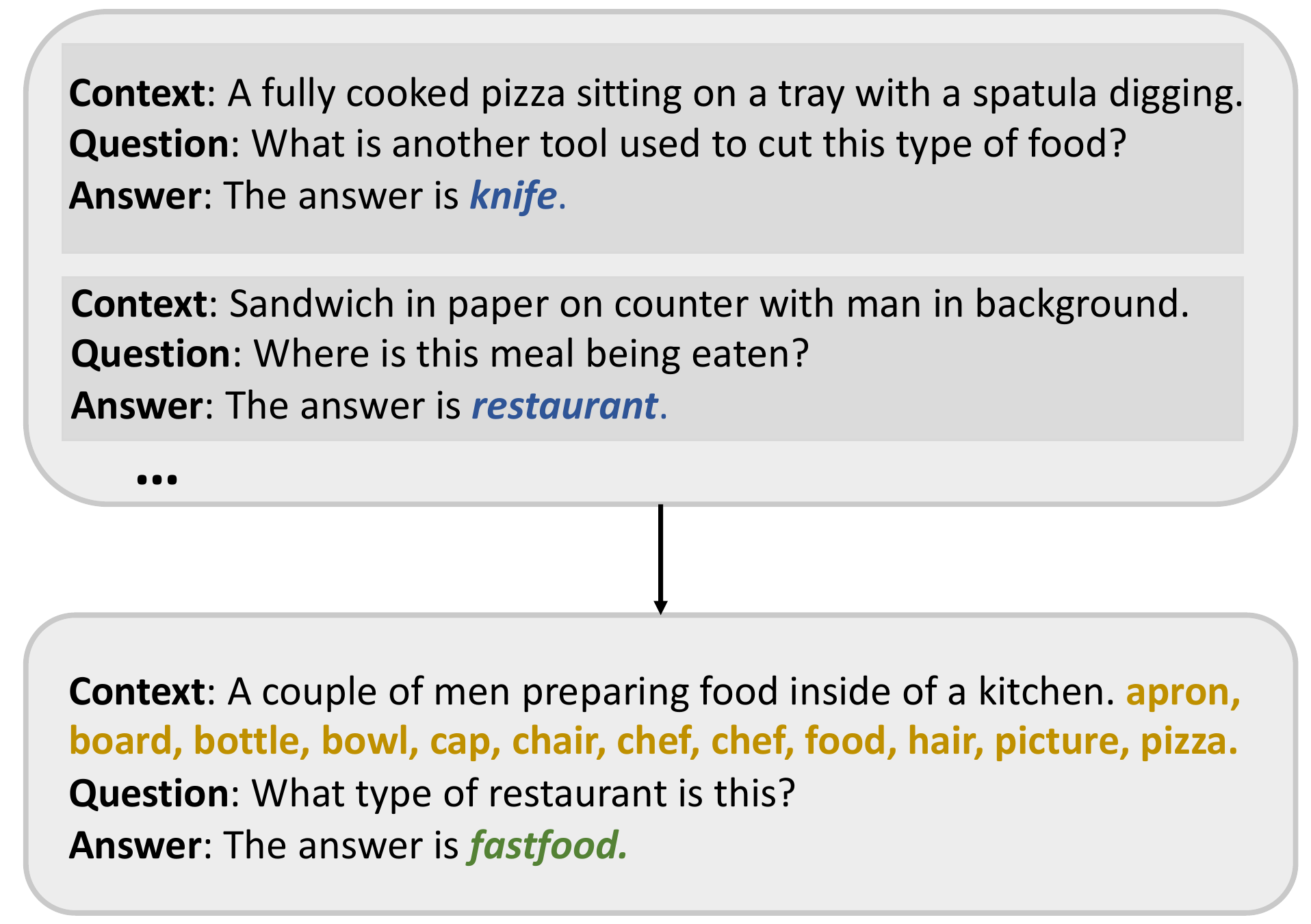}
    \caption{Exemplar prompting templates of the PICa~\cite{yang2022empirical} baseline. Outputs in the in-context examples and test examples are marked with \textbf{\textcolor{blue}{blue}} and \textbf{\textcolor{ForestGreen}{green}} colors. Object tags are marked with \textbf{\textcolor{brown}{bronze}}~colors.}
    \label{fig:pica_template}
\end{figure}

\begin{figure}[t]
    \centering
    \includegraphics[width=0.8\linewidth]{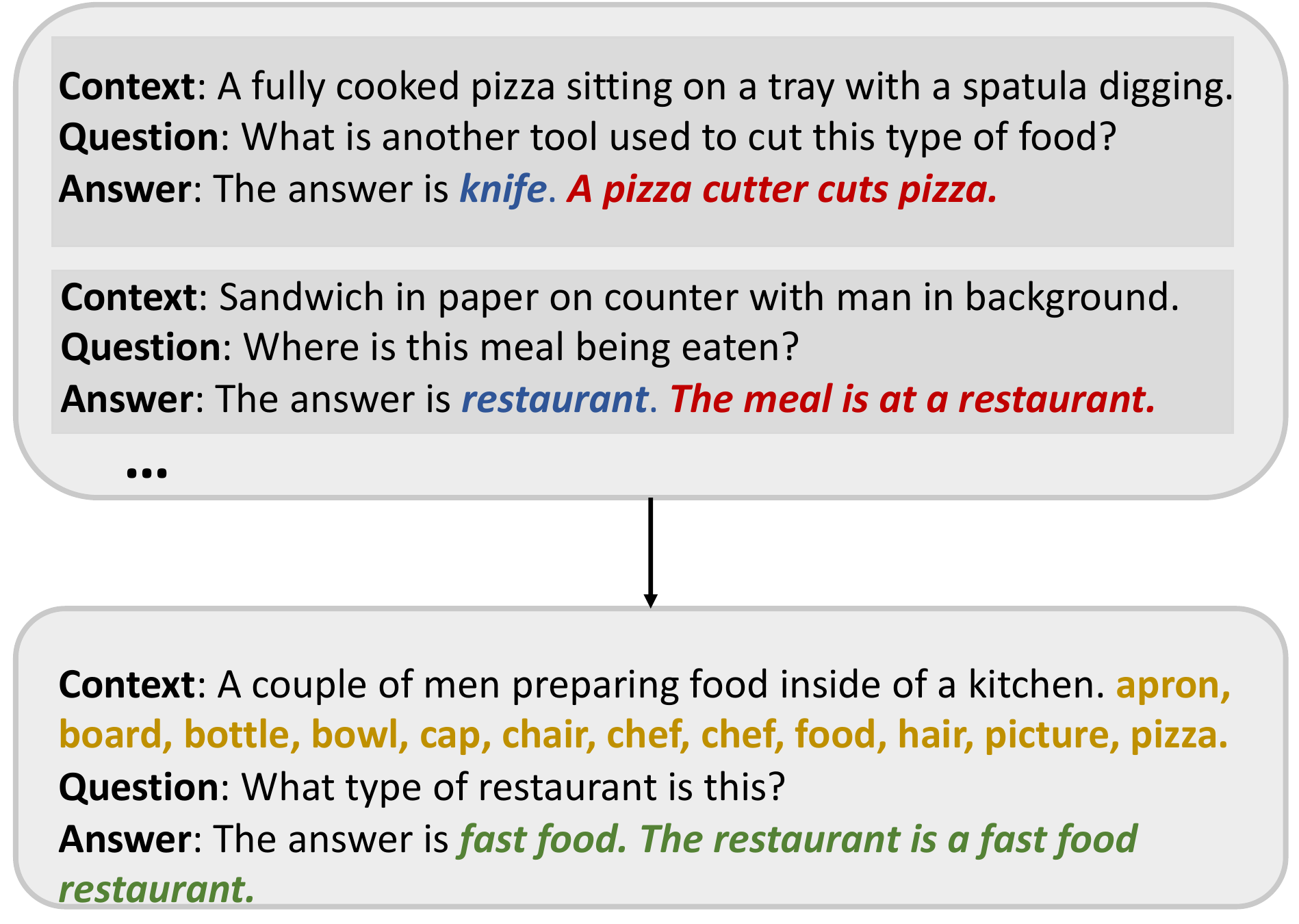}
    \caption{Exemplar prompting templates of the CoT~\cite{wei2022chain} baseline.  Outputs in the in-context examples and test examples are marked with \textbf{\textcolor{blue}{blue}} and \textbf{\textcolor{ForestGreen}{green}} colors. The ``thoughts" (rationales) of the training example is marked with \textbf{\textcolor{red}{red}} colors. Object tags are marked with \textbf{\textcolor{brown}{bronze}}~colors.}
    \label{fig:cot_template}
\end{figure}

\begin{figure*}[t]
    \centering
    \includegraphics[width=\linewidth]{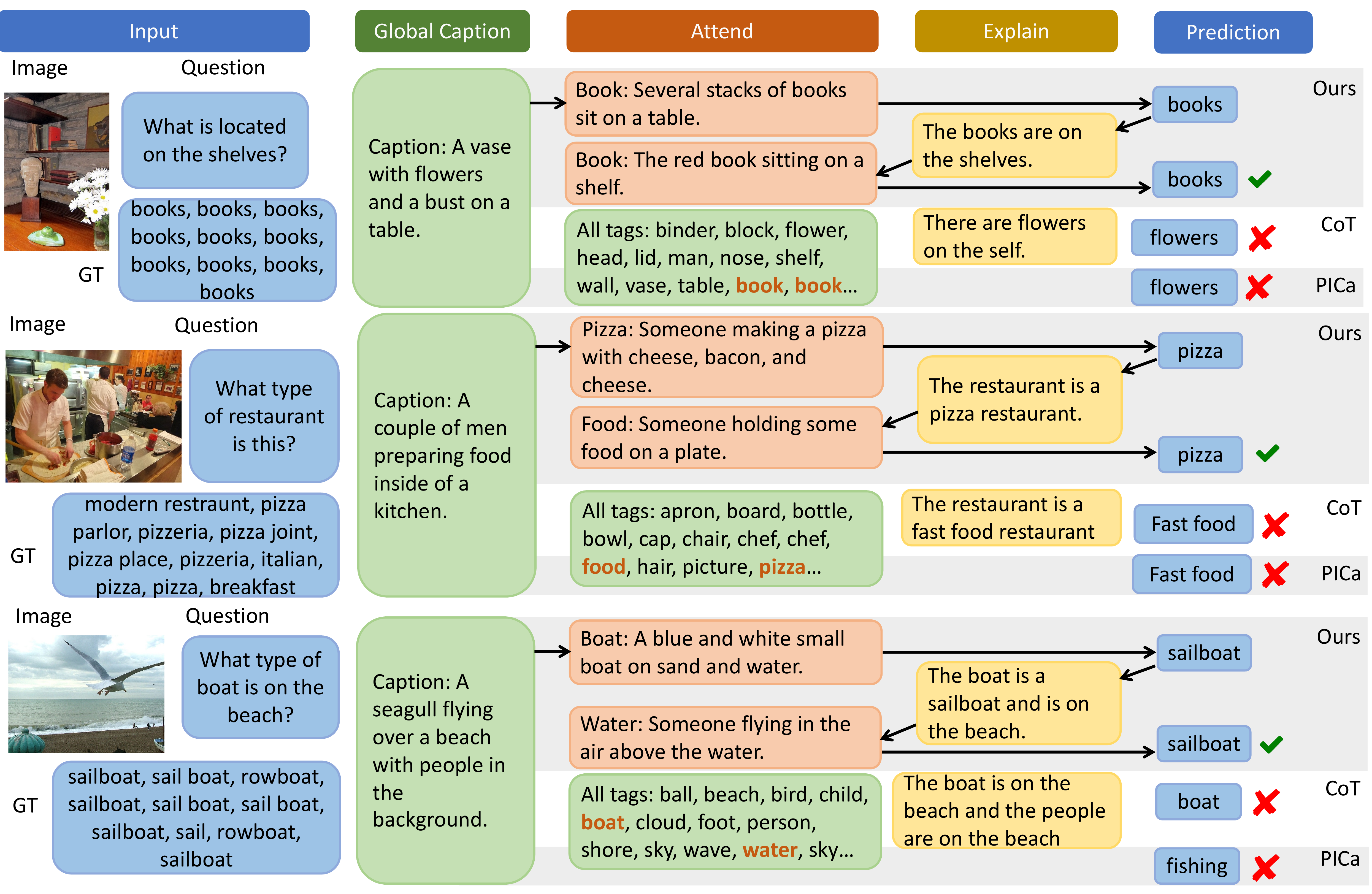}
    \caption{More Qualitative examples of the proposed \model and baselines. Besides better answer accuracy, our method also enjoys better transparency by providing a step-by-step reasoning trace with related visual concepts, regional descriptions, and a supporting rationale. $\rightarrow$ denotes our reasoning flow. }
    \label{sup:qual}
\end{figure*}

\begin{figure*}[t]
    \centering
    \includegraphics[width=\linewidth]{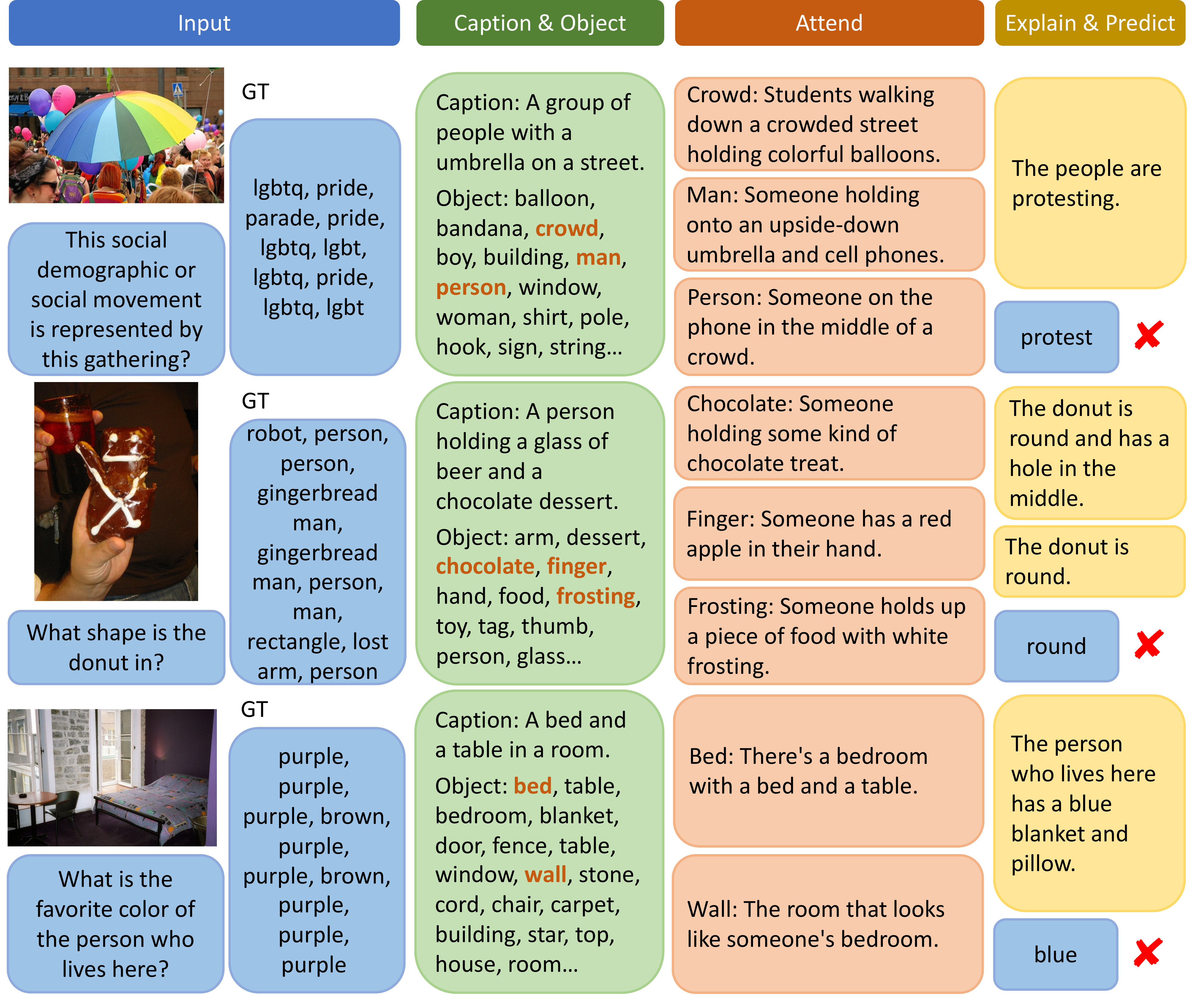}
    \caption{Typical Failure cases of the proposed method and baselines.
    our model provides intermediate results for failure cases. In the first row of the Figure, we show an example that the OPT-66B LLM in our \moduleBB module fails to connect the \textit{``colorful balloons"} and the \textit{``lgbtq"} social movement. In the second row of Figure, we show that our model fails when the vision models fail to capture essential visual context (\ie the shape of the \textit{donut}).
    }
    \label{sup:fail}
\end{figure*}

\end{document}